\documentclass[11pt,a4paper,twoside,openright]{book}
\usepackage[scaled=.95]{helvet}
\usepackage{courier}
\usepackage[T1]{fontenc}
\usepackage[latin1]{inputenc}
\usepackage{graphics, graphicx, epsfig}
\usepackage{times}
\usepackage{epsfig}
\usepackage{graphicx}
\usepackage{amsmath}
\usepackage{amssymb}
\usepackage{dsfont}

\begin{document}

\author{Vincent Garcia and Eric Debreuve and Michel Barlaud}
\title{Fast k Nearest Neighbor Search using GPU}
\maketitle

The recent improvements of graphics processing units (GPU) offer to the computer vision community a powerful processing platform. Indeed, a lot of highly-parallelizable computer vision problems can be significantly accelerated using GPU architecture. Among these algorithms, the $k$ nearest neighbor search (KNN) is a well-known problem linked with many applications such as  classification, estimation of statistical properties, etc. The main drawback of this task lies in its computation burden, as it grows polynomially with the data size. In this paper, we show that the use of the NVIDIA CUDA API accelerates the search for the KNN up to a factor of 120.

\section{Introduction}

A graphics processing unit (also called GPU) is a dedicated graphics rendering device for a personal computer, workstation, or game console. GPU is highly specialized for parallel computing. The recent improvements of GPUs offer a powerful processing platform for both graphics and non-graphics applications. Indeed, a large proportion of computer vision algorithms are parallelizable and can greatly be accelerated using GPU. The use of GPU was, uptil recently, not easy for non-graphics applications. The introduction of the NVIDIA CUDA (Compute Unified Device Architecture) brought, through a C-based API, an easy way to take advantage of the high performance of GPUs for parallel computing.
\\
The $k$ nearest neighbor search problem (KNN) is encountered in many different fields. In statistics, one of the first density estimate~\cite{LoftsgaardenAMS65} was indeed formulated as a $k$ nearest neighbor problem. It has since appeared in many applications such as KNN-based classification~\cite{Dasarathy_1991_BOOK,Shakhnarovich_2006_BOOK} and image filtering~\cite{Yaroslavsky85}. More recently, some effective estimates of high-dimensional statistical measures have been proposed~\cite{Kozachenko_1987_PIT}. These works have some computer vision applications~\cite{Boltz_2006_ICIP, Garcia_2007_ICIP}.
\\
The KNN search is usually slow because it is a heavy process. The computation of the distance between two points requires many basic operations. The resolution of the KNN search polynomially grows with the size of the point sets.
\\
In this paper, we show how GPU can accelerate the process of the KNN search using NVIDIA CUDA. Our CUDA implementation is up to $120$ times faster than a similar C implementation. Moreover, we show that the space dimension has a negligible impact on the computation time for the CUDA implementation contrary to the C implementation. These two improvements allow to (1) decrease the time of computation, (2) reduce the size restriction generally necessary to solve KNN in a reasonable time.

\section{$k$ Nearest Neighbors Search}

\subsection{Problem definition}

\begin{figure}[htbp]
	\centering
	\includegraphics[width=0.7\linewidth]{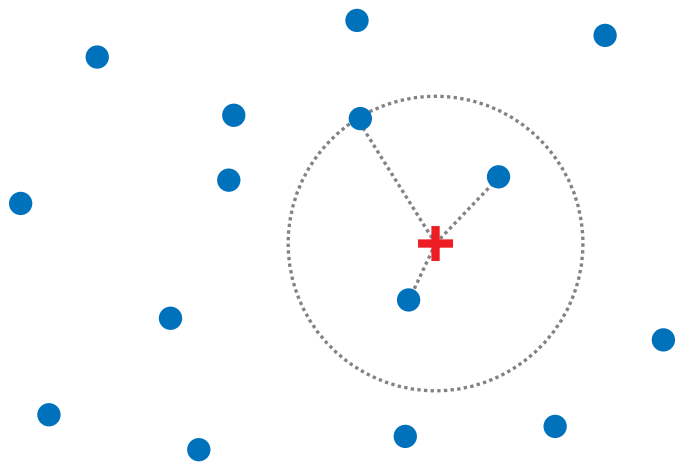}
	\caption{Illustration of the KNN search problem for $k=3$. The blue points correspond to the reference points and the red cross corresponds to the query point. The circle gives the distance between the query point and the third closest reference point.}
	\label{fig:knn_problem}
\end{figure}

Let $R=\{r_1, r_2, \cdots, r_m \}$ be a set of $m$ reference points in a $d$ dimensional space, and let $Q=\{q_1, q_2, \cdots, q_n \}$ be a set of $n$ query points in the same space. The $k$ nearest neighbor search problem consists in searching the $k$ nearest neighbors of each query point $q_i \in Q$ in the reference set $R$ given a specific distance. Commonly, the Euclidean or the Manhattan distance is used but any other distance can be used instead such as infinity norm distance or Mahalanobis distance~\cite{Mahalanobis_1936_NISI}. Figure~\ref{fig:knn_problem} illustrates the KNN problem with $k=3$ and for a set of points in a 2 dimensional space.
\\
One way to search the KNN is the ``brute force'' algorithm (noted BF), also called ``exhaustive search''. For each query point $q_i$, the BF algorithm is the following:
\begin{enumerate}
	\item Compute all the distances between points $q_i$ and $r_j$ with $j$ in $[1,m]$.
	\item Sort the computed distances.
	\item Select the $k$ reference points providing to the smallest distances.
	\item Repeat steps 1. to 3. for all query points.
\end{enumerate}
The main issue of this algorithm is its huge complexity: $O(n m d)$ for the $n m$ distances computed (approximately $2 n m d$ additions/subtractions and $n m d$ multiplications) and $O(n m \log m)$ for the $n$ sorts performed (mean number of comparisons).
\\
Several KNN algorithms have been proposed in order to reduce the computation time. They generally seek to reduce the number of distances computed. For instance, some algorithms~\cite{Arya_1998_ACM} partition the space using a KD-tree~\cite{Bentley_1975_CACM,Indyk_2004_CRCP}, and only compute distances within specific nearby volumes. We show in section~\ref{sec:results} that, according to our experiments, the use of such a method is $3$ times faster than a BF method.
\\
The BF method is by nature highly-parallelizable. Indeed, all the distances can be computed in parallel. Likewise, the $n$ sorts can be done in parallel. This property makes the BF method perfectly suitable for a GPU implementation. According to our experiments, we show in section~\ref{sec:results} that the use of CUDA is $120$ times faster than a similar C implementation and $40$ times faster than a kd-tree based method.

\subsection{Applications}
\label{subsec:applications}

The KNN search is a problem encountered in many graphics and non-graphics applications. Frequently, this problem is the bottleneck of these applications. Therefore, proposing a fast KNN search appears crucial. In this section, we present three important applications using KNN search.
\\

\noindent \textbf{Entropy estimation}\\
In information theory, the Shannon entropy~\cite{Cover_1991_BOOK,Shannon_1948_BSTJ} or information entropy is a measure of the uncertainty associated with a random variable. It quantifies the information contained in a message, usually in bits or bits/symbol. It is the minimum message length necessary to communicate information. This also represents an absolute limit on the best possible lossless compression of any communication: treating a message as a series of symbols, the shortest possible representation to transmit the message is the Shannon entropy in bits/symbol multiplied by the number of symbols in the original message.
\\
The entropy estimation has several applications like tomography~\cite{Gzyl_2002_AMC}, motion estimation~\cite{Boltz_2006_ICIP}, or object tracking~\cite{Garcia_2007_ICIP}.
\\
The Shannon entropy of a random variable $X$ is
\begin{eqnarray}
	H(X) & = & E(I(X)) \\
	     & = & \displaystyle{- \int p(x)\log p(x) dx}
\end{eqnarray}
where $I(X)$ is the information content or self-information of X, which is itself a random variable, and $p$ is the probability density function of $X$.
\\
Given a set of point $Y=\{y_1,y_2,\cdots,y_n\}$ in a $d$-dimensional space, Kozachenko and Leonenko propose in~\cite{Kozachenko_1987_PIT} an entropy estimator based on the distance between each point of the set and its nearest neighbor. Goria \textit{et al.} propose in~\cite{Goria_2005_LMNI} a generalization using the distance, noted $\rho_k(y_i)$, between $y_i$ and its $k$-th nearest neighbor.
\\
The estimated entropy $\widehat{H}_{n,k}(Y)$ depending on $n$ and $k$ is given by
\begin{eqnarray}
	\nonumber \widehat{H}_{n,k}(Y) & = \displaystyle{\frac{1}{n} \sum_{i=1}^{n}} & \big[ \log((n-1) \rho_k(y_i))\\
	& & + \log(c_1(d)) - \Psi(k) \big]
	\label{eq:entropie_knn}
\end{eqnarray}
where $\Psi(k)$ is the digamma function
\begin{equation}
	\Psi(k) = \dfrac{\Gamma'(k)}{\Gamma(k)} = \int_0^\infty \left[ \dfrac{e^{-t}}{t} - \dfrac{e^{-kt}}{(1-e^{-t})}\right] dt
	\label{eq:digamma}
\end{equation}
and
\begin{equation}
	c_1(d) = \dfrac{2 \pi^{\frac{d}{2}}}{d \Gamma(\frac{d}{2})}
	\label{eq:boule}
\end{equation}
gives the volume of the unit ball in $\mathds{R}^d$.
\\

\noindent \textbf{Classification and clustering}\\
The classification is the act of organizing a dataset by classes such as color, age, location, etc. Given a training dataset (previously called reference set) where each item belongs to a class, statistical classification is a procedure in which a class presented in the training dataset is assigned to each item of a given query dataset.
\\
For each item of the query dataset, the classification based on KNN~\cite{Dasarathy_1991_BOOK,Shakhnarovich_2006_BOOK} locates the $k$ closest members (KNN), generally using the Euclidean distance, of the training dataset. The category mostly represented by the $k$ closest members is assigned to the considered item in the query dataset because it is statistically the most probable category for this item. Of course, the computing time goes up as $k$ goes up, but the advantage is that higher values of $k$ provide smoothing that reduces vulnerability to noise in the training data. In practical applications, typically, $k$ is in units or tens rather than in hundreds or thousands. 
\\
The term ``classification'' is synonymous with what is commonly known (in machine learning) as clustering. Statistical classification algorithms are typically used in pattern recognition systems.
\\

\noindent \textbf{Content-based image retrieval}\\
Content-based image retrieval (CBIR)~\cite{Lew_2006_ACM,Lowe_2004_IJCV} is the application of computer vision to the image retrieval problem, that is, the problem of searching for digital images in large databases. ``Content-based'' means that the search will analyze the actual contents of the image. The term ``content'' in this context might refer colors, shapes, textures, or any other information that can be derived from the image itself. The techniques, tools, and algorithms that are used originate from fields such as statistics, pattern recognition, signal processing, and computer vision.
\\
Given an image database and a query image, Schmid and Mohr propose in~\cite{Schmid_1996_ICCV} a simple KNN-based CBIR algorithm:
\begin{enumerate}
	\item Extract keypoints~\cite{Harris_1988,MikolajczykSchmid_IJCV_2004,Schmid_ICCV_1998} in the query image.
	\item Compute the description vector for each extracted keypoint~\cite{Lowe_2004_IJCV,Mikolajczyk_2005_PAMI}. Each vector, also called descriptor, is a set a values describing the local neighborhood of the considered keypoint.
	\item For each descriptor, search in the image database the $k$ closest descriptors according to a distance (typically Euclidean or Mahalanobis distance). Then, a voting algorithm determines the most likely image in the reference image database.
\end{enumerate}
The search of the $k$ closest descriptors is a KNN search problem. The main issue of CBIR is the computation time. In his context, the descriptor size is generally restricted to insure a reasonable computational time. A typical value is between $9$ and $128$.
\\

\section{Experiments}
\label{sec:results}

The initial goal of our work is to speed up the KNN search process in a Matlab program. In order to speed up computations, Matlab allows to use external C functions (Mex functions). Likewise, a recent Matlab plug-in allows to use external CUDA functions. In this section, we show, through a computation time comparison, that CUDA greatly accelerates the KNN search process. We compare three different implementations of the BF method and one method based on kd-tree (KDT)~\cite{Arya_1998_ACM}:
\begin{itemize} 
\item BF method implemented in Matlab (noted BF-Matlab)
\item BF method implemented in C (noted BF-C)
\item BF method implemented in CUDA (noted BF-CUDA)
\item KDT method implemented in C (noted KDT-C)
\end{itemize}
The KDT method used is the ANN C library~\cite{Arya_1998_ACM}. This method is commonly used because it is faster than a BF method. The computer used to do this comparison is a Pentium 4 3.4 GHz with 2GB of DDR memory and a NVIDIA GeForce 8800 GTX graphic card. 
\\
The table~\ref{tab:results} presents the computation time of the KNN search process for each method and implementation listed before. This time depends both on the size of the point sets (reference and query sets) and on the space dimension. For the BF method, the parameter $k$ has not effect on this time. Indeed, the access to any element of a sorted array is done in a constant time. On the contrary, the computation time of the KDT method increases with the parameter $k$. In this paper, $k$ was set to $20$.
\\
\begin{table*}[htbp]
\centering
\begin{tabular}{|c|l|||r|r|r|r|r|r|}
\hline
 & Methods &  N=1200 &  N=2400 &  N=4800 &  N=9600 &  N=19200 &  N=38400 \\
\hline\hline
 D = 8   & BF-Matlab & 0.53 &  1.93 &  8.54 &  37.81 & 154.82 &  681.05 \\
  & BF-C    & 0.55 &  2.30 &  9.73 &  41.35 & 178.32 &  757.29 \\
  & KDT-C   & 0.15 &  0.33 &  0.81 &   2.43 &   \textbf{6.82} &   \textbf{18.38} \\
  & BF-CUDA & \textbf{0.02} &  \textbf{0.10} &  \textbf{0.38} &   \textbf{1.71} &   7.93 &   31.41 \\
\hline
 D=16  & BF-Matlab & 0.56 &  2.34 &  9.62 & 53.64 &  222.81 &  930.93 \\
 & BF-C    & 0.64 &  2.70 & 11.31 & 47.73 &  205.51 &  871.94 \\
 & KDT-C   & 0.28 &  1.06 &  5.04 & 23.97 &   91.33 &  319.01 \\
 & BF-CUDA & \textbf{0.02} &  \textbf{0.10} &  \textbf{0.38} &  \textbf{1.78} &    \textbf{7.98} &   \textbf{31.31} \\
\hline
 D=32  & BF-Matlab & 1.21 &  3.91 & 21.24 & 87.20 &  359.25 & 1446.36 \\
 & BF-C    & 0.89 &  3.68 & 15.54 & 65.48 &  286.74 & 1154.05 \\
 & KDT-C   & 0.43 &  1.78 &  9.21 & 39.37 &  166.98 &  688.55 \\
 & BF-CUDA & \textbf{0.02} &  \textbf{0.11} &  \textbf{0.40} &  \textbf{1.81} &    \textbf{8.35} &   \textbf{33.40} \\
\hline
 D=64  & BF-Matlab & 1.50 &  9.45 & 38.70 & 153.47 & 626.60 & 2521.50 \\
 & BF-C    & 2.14 &  8.54 & 36.11 & 145.83 & 587.26 & 2363.61 \\
 & KDT-C   & 0.78 &  3.56 & 14.66 &  59.28 & 242.98 & 1008.84 \\
 & BF-CUDA & \textbf{0.03} &  \textbf{0.12} &  \textbf{0.44} &   \textbf{2.00} &   \textbf{9.52} &   \textbf{37.61} \\
\hline
  D=80  & BF-Matlab & 1.81 & 11.72 & 47.56 & 189.25 & 761.09 & 3053.40 \\
 & BF-C    & 2.57 & 10.20 & 42.48 & 177.36 & 708.29 & 2811.92 \\
 & KDT-C   & 0.98 &  4.29 & 17.22 &  71.43 & 302.44 & 1176.39 \\
 & BF-CUDA & \textbf{0.03} &  \textbf{0.12} &  \textbf{0.46} &   \textbf{2.05} &   \textbf{9.93} &   \textbf{39.98} \\
\hline
 D=96  & BF-Matlab & 2.25 & 14.09 & 56.68 & 230.40 & 979.44 & 3652.78 \\
 & BF-C    & 2.97    & 12.47 & 49.06 & 213.19 & 872.31 & 3369.34 \\
 & KDT-C   & 1.20 &  4.96 & 19.68 &  82.45 & 339.81 & 1334.35 \\
 & BF-CUDA & \textbf{0.03} &  \textbf{0.13} &  \textbf{0.48} &   \textbf{2.07} &  \textbf{10.41} &   \textbf{43.74} \\
\hline
\end{tabular}
\caption{Comparison of the computation time, given in seconds, of the methods (in each cell respectively for top to bottom) BF-Matlab, BF-C, KDT-C, and BF-CUDA. BF-CUDA is up to $120$ times faster than BF-Matlab, $100$ times faster than BF-C, and $40$ times faster than KDT-C.}
\label{tab:results}
\end{table*}
In the table~\ref{tab:results}, $N$ corresponds to the number of reference and query points, and $D$ corresponds to the space dimension. The computation time given in seconds, corresponds respectively to the methods BF-Matlab, BF-C, KDT-C, and BF-CUDA. The chosen values for $N$ and $D$ are typical values that can be found in papers using the KNN search. 
\\

The main result of this paper is that, in most of cases, CUDA allows to greatly reduce the time needed to resolve the KNN search problem. BF-CUDA is up to $120$ times faster than BF-Matlab, $100$ times faster than BF-C, and $40$ times faster than KDT-C. For instance, with $38400$ reference and query points in a $96$ dimensional space, the computation time is approximately one hour for BF-Matlab and BF-C, $20$ minutes for the KDT-C, and only $43$ seconds for the BF-CUDA. The considerable speed up we obtain comes from the highly-parallelizable property of the BF method.
\\

\begin{figure}[htbp]
	\centering
	\includegraphics[width=0.99\linewidth]{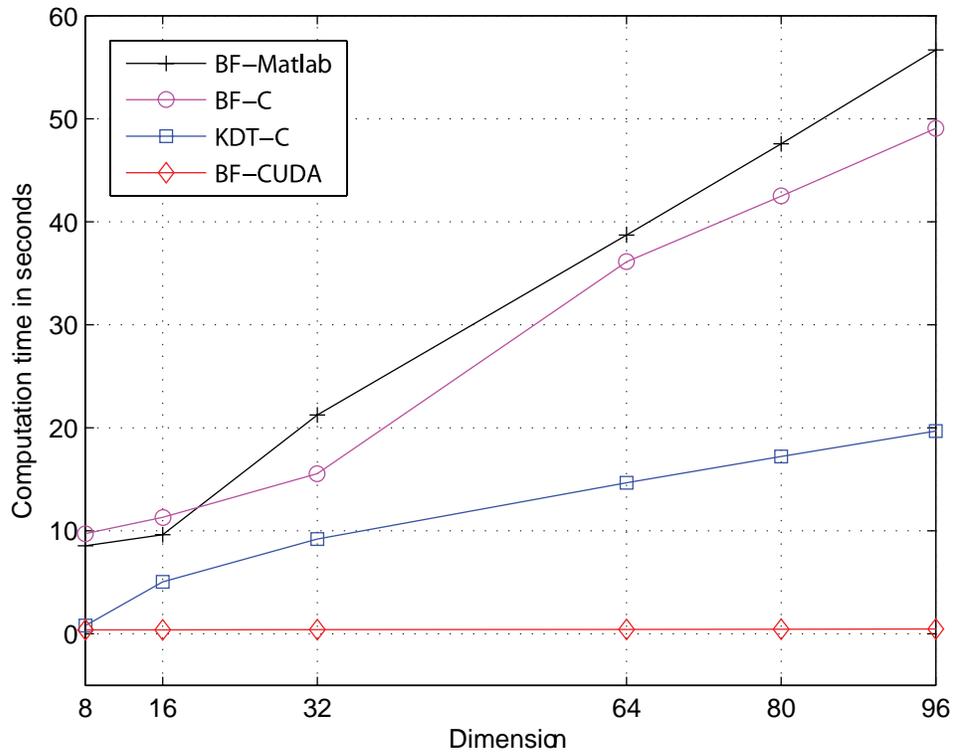}
	\caption{Evolution of the computation time as a function of the point dimension for methods BF-Matlab, BF-C, BF-CUDA, and KDT-C. The computation time increases linearly with the dimension of the points whatever the method used. However, the increase is quasi-null with the BF-CUDA.}
	\label{fig:time}
\end{figure}

The table~\ref{tab:results} reveals another important result. Let us consider the case where $N=4800$. The computation time seems to increase linearly with the dimension of the points (see figure~\ref{fig:time}). The major difference between these methods is the slope of the increase. Indeed, the slope is approximately $0.56$ for BF-Matlab method, $0.48$ for BF-C method, $0.20$ for KDT-C method, and quasi-null (actually $0.001$) for BF-CUDA method. In other words, the methods BF-Malab, BF-C, and KDT-C are all sensitive to the space dimension in term of computation time (KDT method is less sensitive than BF methods). On the contrary, the space dimension has a negligible impact on the computation time for the CUDA implementation. This behavior is more important for $N=38400$. In this case, the slope is $34$ for BF-C, $31$ for BF-Matlab, $14$ for KDT-C, and $0.14$ for BF-CUDA. This characteristic is particularly useful for applications like KNN-based content-based image retrieval (see section~\ref{subsec:applications}): the descriptor size is generally limited to allow a fast retrieval process. With our CUDA implementation, this size can be much higher bringing more precision to the local description and consequently to the retrieval process.
\\

The table~\ref{tab:results} provides further interesting results. First, we said before that, in most of cases, BF-CUDA is the fastest method to resolve the KNN search problem. Let us now consider the cases where $D=8$ and $N=19200$ or $N=38400$. In these cases, the fastest method is the KDT-C. The explanation of why BF-CUDA is not the fastest method is inherent in CUDA. With $D=8$, there are few operations needed to compute the distance between two points and the most of the time is spent in data copies between CPU memory and GPU memory (according to the CUDA profiler). On the contrary, KDT-C does not require this data transfer. With $D>8$, even if the most of the computation time is still spent in memory transfer, BF-CUDA becomes the most interesting implementation.
\\
This table shows also that the KDT implementation is generally $3$ times faster than BF implementation.

\section{Conclusion}

In this paper, we propose a fast $k$ nearest neighbors search (KNN) implementation using a graphics processing units (GPU). We show that the use of the NVIDIA CUDA API accelerates the resolution of KNN up to a factor of 120. In particular, this improvement allows  to reduce the size restriction generally necessary to search KNN in a reasonable time in KNN-based content-based image retrieval applications.

\bibliographystyle{alpha-fr}
\bibliography{main}

\end{document}